\documentclass[12pt]{article}

\usepackage{graphicx}

\usepackage{amsmath}
\usepackage{amssymb}
\usepackage{amsthm}
\usepackage{authblk}

\usepackage[ruled,linesnumbered]{algorithm2e}

\SetAlFnt{\small}
\SetAlCapFnt{\small}
\SetAlCapNameFnt{\small}
\SetAlCapHSkip{0pt}
\IncMargin{-\parindent}
\usepackage{float}
\usepackage[hidelinks]{hyperref}
\hypersetup{
    pdftitle={Lin-DBSCAN},
    bookmarksnumbered=true,     
    bookmarksopen=true,         
    bookmarksopenlevel=1,       
    colorlinks=false,            
    pdfstartview=Fit,           
    pdfpagelayout=TwoPageRight
}

\newcommand{\myalgocapitals}{Lin-DBSCAN }
\newcommand{\myalgocapitalsnospace}{Lin-DBSCAN}
\newcommand{\myalgoextended}{Linear DBSCAN }
\newcommand{\myalgoextendednospace}{Linear DBSCAN}

\newtheorem{definition}{Definition}

\title{Linear density-based clustering with a discrete density model}

\begin{document}



\author[1]{Roberto Pirrone}
\author[1]{Vincenzo Cannella}
\author[ ]{Gabriella Giordano}
\author[ ]{Sergio Monteleone}
\affil[1]{Dipartimento dell'Innovazione Industriale e Digitale (DIID)\\
		Universit\`a degli Studi di Palermo\\
		Viale delle Scienze, Edificio 6, 90128 Palermo, Italy}

\affil[ ]{\textit {\{roberto.pirrone@unipa.it\\ vcannella@gmail.com\\ gabriella.giordano@live.it\\ sergio.monteleone@unipa.it\}}}

\maketitle

%
%
%



\begin{abstract}
Density-based clustering techniques are used in a wide range of data mining applications. One of their most attractive features consists in not making use of prior knowledge of the number of clusters that a dataset contains along with their shape.
In this paper we propose a new algorithm named \myalgoextended (\myalgocapitalsnospace), a simple approach to clustering inspired by the density model introduced with the well known algorithm DBSCAN. Designed to minimize the computational cost of density based clustering on geospatial data, \myalgocapitals features a linear time complexity that makes it suitable for real-time applications on low-resource devices. \myalgocapitals uses a discrete version of the density model of DBSCAN that takes advantage of a grid-based scan and merge approach. The name of the algorithm stems exactly from its main features outlined above. The algorithm was tested with well known data sets. Experimental results prove the efficiency and the validity of this approach over DBSCAN in the context of spatial data clustering, enabling the use of a density-based clustering technique on large datasets with low computational cost.
\end{abstract}




\section{Introduction}
\label{introduction}
In the recent years the research field on efficient clustering techniques gained a major role thanks to the exponential trend of the big-data analytics world. 

Clustering algorithms are used to identify groups of elements that share some distinctive feature. 
The possibility to outline classes of similar elements is extremely relevant in data-mining applications, and most of the times enables further elaboration steps. 

Distinct approaches to clustering produced several algorithms that address the problem of classification differently.

K-means \cite{KMEANS} and several derived variants like K-medoids \cite{k-medoids} are examples of centroid-based clustering methods that use a set of $K$ elements, each one representative of a distinct cluster, to aggregate the others.  
CURE \cite{CURE} and Chameleon \cite{Chameleon} and several others fall into the category of connectivity-based clustering whose peculiarity is the production of multiple partitions of the same dataset that are related in a hierarchical way. 
CLARANS \cite{CLARANS} is an example of partitioning clustering, while BIRCH \cite{BIRCH} is grid-based clustering approach and STING \cite{STING} uses statistical information to cluster data.

DBSCAN (Density-Based Spatial Clustering of Applications with Noise) \cite{DBSCAN} is a milestone in the field of clustering techniques that gave origin to a prolific research trend thanks to its innovative theoretical clustering model based on the concept of density. 
The density-based clustering model consists in a set of topological constraints that provide the aggregation rules to identify clusters.

According to these rules, a point can be part of a cluster if at least a minimum number of elements are included in its neighborhood of radius $epsilon$.
In the rest of this paper we will refer to these parameters as $MinPts$ and $Eps$ respectively, adopting the naming convention used by Ester et al.

The major break-through deriving from the application of this technique, consists in the detection of arbitrarily-shaped isodensity clusters, combined with the contextual discard of outliers, regarded as noise. 

Nevertheless, the price for these remarkable features is the $O(n^2)$ computational complexity, where $n$ equals the cardinality of the input dataset. 
Recent research studies on DBSCAN prove indeed that the complexity of the algorithm depends on the dimensionality $d$ of the dataset, and for $d \ge 3$ it is bound to $\Theta(n^{4/3})$ \cite{Gan:2015:DRM:2723372.2737792}.

We propose here a new clustering algorithm based on an approximated density model derived from the topological definition of cluster that represents the theoretical foundation of DBSCAN. 

The reliability of this approximated density model depends on the discretization of the space that contains the input dataset. The result of this process is an indexing structure whose elements have a regular shape, a uniform extension in space and a set of univocally determined index coordinates.

For 2D and 3D datasets this indexing structure assumes the form of a sparse grid that fills the same region of the original dataset.

Differently from other grid-based density clustering techniques though, \myalgocapitals does not rely on this grid for indexing.
Instead, it uses the number of points that fall within the boundaries of each element as the only criteria to generate clusters on connected elements of the grid, reaching linear computational complexity on geospatial domains.

Obviously, the results produced by \myalgocapitals differ from those of DBSCAN. Nevertheless the difference is negligible for both real-time applications and low-power devices where the performance trade-off is unavoidable. 

As an example, this is the case of computer vision where a large amount of low-dimensional features are extracted with high frame rate from visual perception streams, and they have to be processed in real-time to assess the emergence of relevant patterns for the task at hand.

The rest of this paper is arranged as follows: section \ref{related_works} provides a brief overview on many works that focus on the improvement of both performance and efficacy of DBSCAN, pursuing different approaches. Section \ref{approx_density_model} presents our discrete density model and section \ref{main_algorithm} presents the proposed \myalgocapitals algorithm, focusing on asymptotic complexity analysis, heuristics applied for parameters estimation and parameters sensitivity. 
Clustering evaluation of Lin-DBSCAN is exploited in section \ref{validation}; the reported results show that \myalgocapitals performs better than DBSCAN according to both internal and external indexes, when running on several benchmark clustering datasets.
Section \ref{experiments} presents the results achieved using this algorithm with some well-known benchmark datasets. Section \ref{parallel} reports some hints for a parallel implementation. 
Section \ref{application} discusses the application of \myalgocapitals in a real-time robot vision task, comparing the results to the original implementation that made use of DBSCAN.
Finally, section \ref{conclusions} presents our conclusions and future works.

\section{Related works}
\label{related_works}
Several efforts were devoted in finding solution to improve the efficiency of DBSCAN.
The first optimizations consisted in alternative implementations that made use of spatial indexing structures (R*-trees \cite{RStarTree} or KD-trees \cite{Volker}) to reduce the average computational cost to $O(n\log(n))$. 

Other variants focused on optimizing the cost of the neighbors search following various strategies, resulting in purely density-based or hybrid variants. 
 
IDBSCAN (Improved sampling-based DBSCAN) \cite{IDBSCAN}, samples the neighborhood of a given point, focusing only on so-called seed points that lie on the $Eps$-radius circumference. This sampling technique reduces the subsequent neighbors search, but the algorithm still preserves an $O(n\log(n))$ computational complexity. 
 
FDBSCAN (Fast Density-Based Clustering for Applications with Noise) \cite{FDBSCAN} sorts the input data by an arbitrarily chosen dimensional coordinate. This pre-processing phase lowers the cost for the clustering step to an approximately linear time complexity. However the additional cost for the preliminary sorting step bounds the overall cost to $O(n \log(n))$.

TI-DBSCAN (Triangular Inequality DBSCAN) \cite{TIDBSCAN} avoids the use of indexing structures, in favor of the triangular inequality property to reduce the neighborhood search space even in high-dimensional domains.

Other variants try to overcome the limits of DBSCAN in detecting clusters with different densities.

As an example, LDBSCAN (Local-Density Based Spatial Clustering for Applications with Noise) \cite{LDBSCAN} uses a slightly modified model that provides the definition of local density to identify non-isodensity clusters.

OPTICS (Ordering Points To Identify the Clustering Structure) \cite{OPTICS} is a hybrid approach that exhibits features common to both the density and hierarchical clustering paradigms. The output is a special kind of dendrogram of the elements of the dataset, called \textit{reachability-plot}. All the elements that fall into valleys of the plot belong to the same cluster. 

HDBSCAN (Hierarchical DBSCAN) \cite{HDBSCAN} is a hybrid hierarchical and density-based technique, that similarly to OPTICS produces a dendrogram and then applies a further step to ``condense'' the dendrogram tree into a real set of clusters.

In \cite{MAHESHKUMAR201639} the Group graph-based index structure is proposed to overcome the problems of hierarchical indexing that fails to scale for datasets of dimensionality above 20. Authors report a performance increase of a factor 1.5-2.2 on benchmark datasets.

Several other versions try to address both the problems of complexity and varying density integrating grid-based clustering techniques.

GF-DBSCAN (Grid Fast density-based clustering for Applications with Noise) \cite{GFDBSCAN} partitions the input space into $Eps$- sized cells. The grid-partition is used to confine the the neighbors search of each point to the containing cell or in the near ones. A check based on distance evaluation is required in order to detect the neighbors for each point. Better performances than FDBSCAN have been reported.

GRIDBSCAN (GRId Density-Based Spatial Clustering of Applications with Noise)\cite{GRIDBSCAN1} uses a regular grid to cope with varying density. The algorithm produces an estimation of the $Eps$ parameter for each region of the grid that exhibit a similar density and then resort to DBSCAN to produce the clustering results. 

GriDBSCAN \cite{GRIDBSCAN2}, as GF-DBSCAN, uses a regular grid, whose cells have $Eps$-sized sides. DBSCAN is carried out in each cell, and the clusters belonging to different cells are subsequently merged if needed.

Very recently, GridDBSCAN \cite{Kumari:2017:EFS:3007748.3007773} has been proposed that makes use of Grid-R-tree, a modified version of the R-tree algorithm, to make grid-wise queries efficient. Authors claim to run up to 8.98x faster than $O(n \log{n})$ grid-based approaches, and compare the shared and distributed memory versions of GridDBSCAN against the analogus versions of PDSDBSCAN \cite{Patwary:2012:NSP:2477183.2478122}, the well-known 2012 approach by Patwary et al.

In the last few years, several algorithms suitable for parallel and distributed architectures have been proposed; some examples of this category are PDBSCAN (Parallel-DBSCAN) \cite{ParallelDBSCAN}, Mr. Scan \cite{MrScan}, Pardicle \cite{Pardicle}, MR-DBSCAN \cite{MRDBSCAN}, HPDBSCAN \cite{Gotz:2015:HHP:2834892.2834894}, MRG-DBSCAN \cite{ma2015mrg}, RDD-DBSCAN \cite{DBLP:conf/ieeehpcs/CordovaM15}, BD-CATS \cite{BDCATS}, and the parallel approach presented in \cite{7536435}, that uses MPI on different threads running classical DBSCAN on suitable data subsets. A similar approach is adopted by S\_DBSCAN \cite{7723739}, which relies on the Apache Spark framework to support the initial data partition based on random sample, manage parallel threads running DBSCAN on each partition, and merging the results based on the centroid position in each cluster.

In general, these DBSCAN variants inspired to parallel programming focus on scaling the applicability of DBSCAN to big-data domains preserving the density-based clustering benefits. Although the use of parallel processors and distributed architectures reduces significantly the computational costs, in many cases these are not feasible solutions for real-time applications on low-power devices, that for energy-balance reasons are not equipped with a sufficiently high number of core processors. 

Aside from big-data domains, the appeal of a density-based clustering is also strong in many applications of computer vision \cite{lanedetect}\cite{7838237}, pattern recognition \cite{Cao_2015_CVPR}, real-time video \cite{crowddetect} and superpixel \cite{Shen:2016:RSS:3012252.3026622} segmentation etc., where clustering is a crucial pre-processing step to enable discovering interesting patterns related to the task under investigation.

Our contribution focuses on the design of highly efficient clustering techniques for real-time applications on spatial domains, whose applicability can be extended to low-power devices. 

We developed a clustering algorithm named \myalgoextended (\myalgocapitalsnospace), that is based on an approximation of the clustering model first introduced with DBSCAN. This new algorithm relies on space discretization to increase the efficiency up to linear computational cost on low-dimensional datasets. 

\section{Motivations: approximation of density-based clustering by space discretization}
\label{approx_density_model}
The most critical issue of DBSCAN, and of many of its variants, is the need to perform several neighborhood search operations. 

This routine implies a high computational cost, proportional to the cardinality of the dataset. In the previous section, we reviewed several variants of the original DBSCAN algorithm that improve the execution time, but are, however, bound to an $O(n \log{n})$ complexity. 

\myalgocapitals was designed to overcome the limitations of DBSCAN in terms of computational complexity on low-dimensional datasets, without necessarily crossing the boundaries of sequential programming.
In order to achieve a linear time complexity, \myalgocapitals implements a hybrid model that integrates features from both grid and density-based clustering techniques.

We call the theoretical foundation on which \myalgocapitals is based, \textit{discrete density-based model}. It can be considered as an approximation of the density model used by DBSCAN, that takes advantage of a uniform discretization of dense regions of the domain space. 

A grid, whose partition matches this uniform discretization step, is used to index the input points before proceeding with the evaluation of the density-connected regions. 
A similar approach can be found in \cite{Lasek2009} and \cite{Lasek2013}.

However, unlike other grid-based algorithms, in \myalgocapitals the grid is not simply used as an indexing structure but is, indeed, the result of a transformation of the input dataset, which allows the algorithm to work on a simplified domain; i.e. the detection of density-connected regions operates on grid cells directly, rather than evaluating each point and its neighborhood.

\subsection{Selection of the discretization step}
\label{discretization_step}

Similarly to other grid-based clustering techniques, we investigate the presence of clusters of elements on the hyperrectangle that encloses all the elements of the input dataset.
\myalgocapitals operates a uniform partitioniong on this hyperrectangle by the superimposition of a multidimensional grid whose step matches the selected discretization step. 
Therefore, each element of the grid is a hypercube whose edges have all the same length, that is equal to the grid step.
The following definitions provide the conceptual grounding for the selection of an appropriate value of the discretization step.

\begin{definition} [Cell]
\label{cell_def}
A cell is an element of the grid univocally identified by the set of its indices, one for every dimension of the domain space indexed by the grid.
\end{definition}

\begin{definition} [Cell size]
\label{cell_size_def}
The size of a cell is the length of its edge.
\end{definition}

\begin{definition} [Cell cardinality]
\label{cell_cardinality_def}
The cardinality of a cell equals to the number of points that it contains. We will use the notation $|c|$ to indicate the cardinality of a cell.
\end{definition}

The uniform discretization step used by \myalgocapitals is directly related to the $Eps$ parameter of DBSCAN. 
We call this value $\gamma$ and use it as the cell size of the grid used by \myalgocapitalsnospace: 
\[
\gamma = \frac{Eps}{2 \times \sqrt{2}}
\] 

\begin{figure}[H]
	\centering
	\frame{\includegraphics[width=0.32\textwidth]{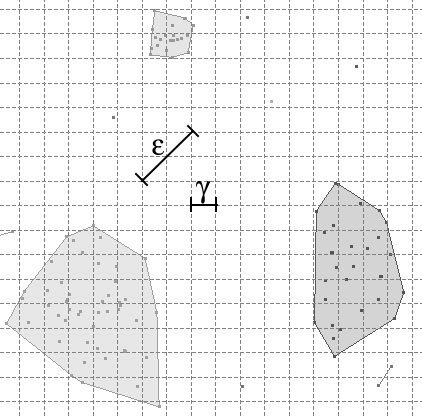}}
	\caption{Relation between $Eps \; (\epsilon)$ and $\gamma$ on a bi-dimensional dataset. The cells of the grid have a square shape with the length of the edge equal to $\gamma$.}
	\label{fig:grid}
\end{figure}

Figure \ref{fig:grid} shows the relation between $Eps$ and $\gamma$ for a bi-dimensional dataset.
Making cells size smaller than $Eps$ ensures that all the points falling inside the same cell are part of the same cluster if the cardinality of the cell is at least $MinPts$.

\subsection{Aggregation rules}
\label{aggregation_rules}
The set of constrains used for the detection of clusters is given by the following definitions that, jointly, constitute the aggregation rules used by \myalgocapitalsnospace.

\begin{definition} [Neighbors cells]
\label{cell_neigh_def}
$c_1$ is a neighbor cell of $c_2$ if it is comprised in its set of adjacent cells. 
\end{definition}

\begin{figure}[h!]
\label{moore_neighborhood_def} 
	\centering
	\includegraphics[width=0.15\textwidth]{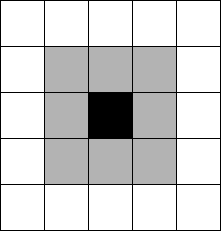}
	\caption{The Moore neighborhood with range $r=1$ of the black cell is composed by the adjacent gray cells.}
	\label{fig:moore_neig}
\end{figure}

The set of adjacent cells is outlined by the definition of Moore neighborhood with range $r=1$, borrowed from the cellular spaces theory \cite{CellularSpaces}. Figure \ref{fig:moore_neig} shows the set of adjacent cells for a two-dimensional grid.

\begin{definition}[Cell-based neighborhood]
\label{cellbased_neigh_def1} 
Two points, $p_1$ and $p_2$, are neighbors if they are located in the same cell $c$ of size $\gamma$: 
\[ 
\forall \; (p_1, p_2) \in c \to p_1 \mbox{ neighbor of } p_2, \ p_2 \mbox{ neighbor of } p_1
\] 
or if they are located in two cells $c_1$ and $c_2$ of size $\gamma$ that are neighbors according to definition \ref{cell_neigh_def}:
\[
\forall \; p_1 \in c_1, \ p_2 \in c_2  \quad p_1 \mbox{ neigh. } p_2, \ p_2 \mbox{ neigh. } p_1 \iff c_1 \mbox{ neigh. } c_2 
\]
\end{definition}

\begin{definition} [Connected cells]
\label{cell_conn_def}
A cell $x$ is connected to a cell $y$ if exists a set of cells $\{c_i \,\, | \,\, \forall i$ $c_i \, \text{is a neighbor of} \, c_{i+1}\}$, where $i=1,2,..,N$ and $c_1 = x, \, c_N = y$. 

We will use the notation $x \leftrightarrow y$ to indicate that two cell are connected. 
\end{definition}

\begin{definition} [Cell-based cluster]
\label{cellbased_cluster_def}
A cluster $C$ is composed of all the points included into a set $S$ of connected cells, each with cardinality greater or equal to $MinPts$:

\[
C = \{p_j\ \, | \,\, p_j \in S\} 
\]
\[
\text{where} \,\,\, S = \bigcup\limits_{i=1}^{N} c_{i} \,\,\, | \,\,\,  \forall i  \,\, |c_i| \ge MinPts
 \,\,\, \text{and} \,\,\, c_i \leftrightarrow c_j \,\,\, \forall j \ne i \,\,\, \text{with} \,\,\, i,j=1,2,..,N 
\]
\end{definition}

Definition \ref{cellbased_neigh_def1} specifies the key aspect of the discrete density model used in \myalgocapitalsnospace: all the points that fall inside the same cell or adjacent cells are also neighbors. 

This is motivated by the fact that, after the association of each point to a cell in the grid, the cell size $\gamma$ guarantees that all the points inside the same cell are neighbors according to the definition of the $Eps-neighborhood$ of DBSCAN.

Moreover, according to definition \ref{cell_conn_def} two points $p$ and $q$, each one falling into a different cell $c_p$ and $c_q$, will be part of the same cluster if and only if $c_p \leftrightarrow c_q$. 
Finally, definition \ref{cellbased_cluster_def} ties the notion of cluster to the concept of set of connected cells denoted by a minimum cardinality.
For bi-dimensional datasets, the way \myalgocapitals detects such regions, resembles the well-known \textit{flood-fill} coloring techniques based on the detection of similarly-colored pixel areas. In this case, instead of the color, the relevant attribute is the minimum density requirement represented by $MinPts$.
The individuation of connected cells can be performed in linear time over the total number of non empty cells in the grid. Moreover, the total number of non-empty cells is typically less than the total number of input points, thus implying an even further reduction of the execution time.

Due to the space discretization, it is obvious that DBSCAN and \myalgocapitals will eventually produce different outputs. However, as shown in the following sections, experimental results prove the validity of this approximated model.

\subsection{Correctness and reliability}
\label{correctness}
\myalgocapitals detects clusters evaluating the connected cells that enclose dense regions of the dataset, without the need to explicitly compute the distances between points. 

Nevertheless, the use of a distance function is implicitly applied in the evaluation of the relation of adjacency between cells.
In order to support the formal correctness in the application of this approximated model, we need to highlight some key aspects on the metric that is used.
Even if the original density-based model of DBSCAN is designed to be independent from the metric, for many applications, the Euclidean distance is often the best choice for elements enclosed in low-dimensional domains, especially with reference to geospatial datasets. 
Furthermore, density-based clustering techniques often prove to be inefficient on high-dimensional spaces and constrain the quality of the results to the selection of an appropriate metric for the domain at hand. 
Therefore, we will grant the correctness of this approximated density model on the application of \myalgocapitals to domains falling into Euclidean two-dimensional and three-dimensional spaces.

We purposely overlook the application of our approximated model to high-dimensional datasets where the density concept itself fails to retain its meaningfulness because of the difficulty in defining valid criteria for the detection of outliers \cite{outliers}, a problem known as \textit{curse of dimensionality}. 
Recent works \cite{HiDimCurse}, suggest that similarity functions other than the Euclidean distance function should be applied in those contexts, and is indeed common that, in high-dimensional domains, the plain density-based approaches are superseded by more specific clustering techniques whose application field falls out of our scope. An interesting survey on this topic is provided in \cite{Kriegel}.   

Another interesting approach to this problem based on the definition of sparse grids can be found in \cite{bungartz_griebel_2004} where an attempt to cope with high dimensionality, also in the context of clustering applications, is presented.

In this section we also want to discuss the reliability of the discrete density-based model of \myalgocapitals with reference to missing data. With the arise of big-data analytics, the problem of missing data became a relevant issue since many clustering algorithms were not designed to handle incomplete datasets.

Following the plain approach of the density model of DBSCAN, \myalgocapitals does not introduce any peculiar technique to handle such cases for the following reasons:

\begin{itemize}
\item missing data is, most of the time, affecting high-dimensionality datasets that, for the reasons given above, are out of the scope of \myalgocapitalsnospace. Furthermore, in such contexts, the reliability of the imputation method used to classify incomplete elements is constrained to the analysis of the domain at hand, the type of missing data and the specific mechanism behind the loss of data \cite{5664691}.
\item for spatial datasets, that are the main use case of our algorithm, several imputation methods are available to resolve the uncertainty in the assignment of an element of the dataset with incomplete features to the cluster with highest affinity \cite{missingdata}. 
\end{itemize}   

We therefore defer the resolution of the missing data issue to one of these alternatives:
\begin{itemize}
  \item a pre-processing step, expressly designed for the application at hand, with the purpose to fill the missing values.
  \item a post-processing step that can assign the incomplete data to the best matching cluster.
\end{itemize}

In both cases, the additional algorithmic step needed to handle incomplete data can be applied with linear computational cost.

\section{The \myalgocapitals algorithm} 
\label{main_algorithm}
The processing flow of \myalgocapitals is composed of two sequential steps that concur in the determination of the final results: the grid creation and the clustering phase.
In this section we will discuss the pseudo-code of the main algorithm, presented in listing \ref{algo:main_algorithm}.
Some details about the inner workings will be discussed next, with the description of two auxiliary functions that are used to provide the final results.

The input dataset, \emph{points}, the cell size $\gamma$, according to definition \ref{cell_size_def}, and the density threshold value $MinPts$ are the input arguments of \myalgocapitalsnospace.
The output is a list that contains all the detected clusters.

The grid is represented in the form of a hash table that associates each cell to a key derived from its set of coordinate indices. 
The empty hash table $Grid$ is initialized at line \ref{algo:grid_init}, and it is then populated scanning all the elements of the input dataset in the loop at line \ref{algo:step1}. 
Note that only non-empty cells will be stored into $Grid$.   
   
\null
\SetAlgorithmName{Listing}{}{}
\begin{algorithm}[H]
\KwIn{points, $\gamma$, $MinPts$}
\KwOut{a list of clusters named $Results$}
  \tcc{Initialization of empty data structures}
  \lnl{algo:grid_init} $Grid$ = $\emptyset$ \;
  \lnl{algo:res_init} $Results$ = $\emptyset$ \;
  
  \tcc{Step 1: grid creation}
  \lnl{algo:step1} \ForEach{$p$ in points}  
  { 
	  $coords$ = floor($p$.coords / $\gamma$) \; 
	  $cell$ = $Grid$.getCell[$coords$] \;
	  \If{$cell$ is null} {
      $cell$ = $Grid$.makeNewCell($coords$) \;
	  }
	  $cell$.addPoint(p) \;
  }

  \tcc{Step 2: clustering}
  \lnl{algo:step2} \While{$Grid$ is not empty} {
    $cell$ = $Grid$.firstCell()\;
    $newCluster$ = makeEmptyCluster()\;
    fill($Grid$, $newCluster$, $cell$, $MinPts$) \tcc*[r]{fill function}
    \If {$newCluster$ is not empty}{
      \lnl{algo:addcluster} $Results$.addCluster($newCluster$)\;
    }
  } 
  \Return{$Results$}
  \caption{the \myalgocapitals algorithm}
  \label{algo:main_algorithm}
\end{algorithm}

\null

The clustering phase starts right after the grid is populated, at line \ref{algo:step2}. 
In this second loop, starting from an arbitrary cell in the grid, the recursive function \emph{fill} aggregates clusters scanning all the neighbors cells.
The selection of the first cell is not affecting the computational complexity of the algorithm, since \myalgocapitals analyzes each cell exactly once. In this case we used the \emph{firstCell} function that returns the first cell in the hash table, but it could be substituted with any function returning a random cell or a null reference if the hash table is empty. 
The same considerations apply to the final results, since the same cluster will be detected starting from any of its component cells, according to definition \ref{cellbased_cluster_def}.

The \emph{fill} function returns when there are no more neighbors cells to explore. If the resulting cluster is not empty it is added to the $Results$ list at line \ref{algo:addcluster}.  
The algorithm stops only when the hash table is empty, i.e. all the cells allocated in the loop starting at line 6 have been scanned.
 
\subsection{The \emph{fill} function} 
\label{fill_function}
Listing \ref{algo:fill} contains the recursive implementation of the \emph{fill} function. 
The first instruction, after an existence check, consists in the immediate removal of the cell from the hash table $Grid$ at line \ref{algo:delcell}. 
 
This step ensures that each cell is evaluated only once during the whole clustering phase.
If the cardinality of the cell is greater or equal than $MinPts$, each point is added to the current cluster (line \ref{algo:addpoints}).
Next, the function \emph{findNeighbors} is invoked at line \ref{algo:expand} to retrieve the list of the neighbors cells still present into the $Grid$ hash table. 

The function \emph{fill} is then recursively invoked on all the neighbors to expand the current cluster. 
The recursion ends when there are no more neighbors cells to scan.

\null
\SetKwProg{FnFill}{Function}{}{return}
\SetAlgorithmName{Listing}{}{}
\begin{algorithm}[H]
  \FnFill{fill ($Grid$, $cluster$, $cell$, $MinPts$)}
  {
    \If{$cell$ is not null}
    {
	    \lnl{algo:delcell} $Grid$.remove($cell$)\;

	    \lnl{algo:minpts} \If{$cell$.numPoints $>= MinPts$\;}
	    {
		    \ForEach{$p$ in $cell$.points}
		    {
			    \lnl{algo:addpoints} cluster.addPoint($p$)\;
		    }				

		    \lnl{algo:expand} $neighborCells$ = findNeighbors($Grid$, $cell$) \tcc*[r]{findNeighbors function}
		    \ForEach{$neighbor$ in $neighborCells$}
		    {
			    \lnl{algo:recursion} fill($Grid$, $cluster$, $neighbor$, $MinPts$) \tcc*[r]{recursion}
		    }
	    }
    }
  }
  \caption{the \emph{fill} function}
  \label{algo:fill}
\end{algorithm}

\null

Two main considerations follow from the analysis of this pseudo-code:
\begin{itemize}
  \item All the elements belonging to cells that do not comply with the minimum density requirement $MinPts$ are discarded as noise points.
  \item The minimum number of elements in a cluster is $MinPts$. This is a consequence of the instruction at line \ref{algo:minpts}. 
\end{itemize} 

\subsection{The \emph{findNeighbors} function}
\label{findNeighbors_function}
Listing \ref{algo:findNeighbors} shows the pseudo-code of the function \emph{findNeighbors} used in \emph{fill}.
For the implementation of this function, the notion of \emph{cell-neighborhood} provided in definition \ref{cell_neigh_def} was used.

\null
\SetKwProg{FnFindNeig}{Function}{}{}
\begin{algorithm}[H]
  \SetAlgorithmName{Listing}{}{}
  \FnFindNeig{findNeighbors($Grid$, $cell$)}
  {
    \lnl{algo:neiCells} $neiCells = \emptyset$\;
    \lnl{algo:ndims} $dims$ = $cell$.coords.length \tcc*[r]{extension to the n-dimensional case}
    \lnl{algo:neiCount} $neiCount$ = pow(3, dims) \tcc*[r]{maximum number of neighbors}
    \lnl{algo:neighbors_loop} \For {k=0 \KwTo $neiCount-1$}
    {
      $neighborCoords = \emptyset$\;

      \lnl{algo:neiloop} \For{d = 0 \KwTo $dims-1$}
      {
        $div$ = pow(3,d)\;
        $offset$ = $(floor ( \frac{k}{div} ) \bmod 3) - 1$ \;
        $neighborCoords$.at(d) = $cell.coord(d) + offset$ \;
      }   
      
      $neighborCell$ = $Grid$.getCell($neighborCoords$)\;
      \If{$neighborCell$ is not null}
      {
        \lnl{algo:neiadd} $neiCells$.append($neighborCell$))\;
      } 
      
    }
  }
  \Return{$neiCells$}
  \caption{the \emph{findNeighbors} function}
  \label{algo:findNeighbors}
\end{algorithm}

\null

The empty list of neighbors cells is initialized at line \ref{algo:neiCells}. 

For the sake of generality, the evaluation of the cell neighborhood is extended to the $n$-dimensional case as a consequence of the instruction at line \ref{algo:ndims}.
Nevertheless, the considerations about the application of this approximated model to high-dimensional domains expressed in section \ref{correctness} remain unchanged.

The maximum number of cells in the neighborhood is then calculated at line \ref{algo:neiCount}.
For simplicity the instructions \ref{algo:ndims} and \ref{algo:neiCount} are included into this listing, but they could be easily removed from the body of this function, since the number of dimensions and the number of neighbors could be calculated once for the whole dataset at the beginning of the algorithm.

The loop at line \ref{algo:neiloop} implements a simple mapping from linear to subscript indices in order to calculate the indices of each neighbors cell from the relative offsets with reference to the indices of the current cell.

The function returns the list of all the existing neighbors at line \ref{algo:neiadd}. 

A reference implementation of \myalgocapitals in C++ can be found at the following URL: \url{https://git.io/v2EFC}.

%
%
%
%

\subsection{Computational Complexity}\label{complexity}
In order to create and populate the grid, \myalgocapitals evaluates each point in the input data set once. For each iteration, it computes the indices of the containing cell, verifying if a non-empty cell already exists in the grid hash map. Using a hash map guarantees that the average cost required to access a cell is constant and therefore, the total cost of this step is $O(n)$ where $n$ is total number of points in the input dataset.

In the clustering phase, the algorithm accesses each cell only once. Therefore, the computational cost of the \emph{fill} procedure is linear with the number of non empty cells in the grid, $C$, that is bound to the distribution of the dataset and to the size of the cell according to definition \ref{cell_size_def}. Empty cells are not stored in the hash map and thus not evaluated.

Therefore the overall computational complexity of the algorithm is: 
\[ O(n) + O(3^d \times C) \]

In the \textbf{optimal} case each input point is contained in the same cell and thus, the overall cost of the algorithm is $O(n) + O(1)$.

In the \textbf{worst} case, the input points are all spread in a different cell, i.e. the number of cells is equal to the number of points. In this case, the total cost is $O(n) + O(3^d \times n)$ where $d$ is the dimensionality of the dataset, and $3^d$ is the number of a cells in a neighborhood with reference to definition \ref{cell_neigh_def}.

In the \textbf{average} case, which is the one of actual interest, the number of non empty cells is far less than the number of elements in the datasets, that is $C << n$, and therefore the exponential factor bound to the dimensionality of the dataset is drastically reduced, especially in the case of geospatial datasets.


With reference to the this last consideration, it is also worth mentioning that density-based approach are inherently bound to the problem known as \textit{curse of dimensionality}, that affects every clustering technique relying on the definition of a distance function in high-dimension domains. 

Therefore the reliability of a density-based clustering approach in high-dimensional domains is in general questionable, and its effectiveness is bound to domain specific features, e.g. the definition of a meaningful distance function.
This consideration implicitly discourages the application of this algorithm to high-dimensional domains, thus legitimate the stress on the linearity of the computational cost. 

Indeed, the design of \myalgocapitals has been aimed to push the applicability of density-based clustering in the emerging context of real-time applications based on the elaboration of raw sensor data on low-power devices equipped with photo, video or depth cameras, or other capture devices for the acquisition of spatial information.

As an example of adherence to this guide line, the algorithm never has to resort to the direct calculation of distances between elements of the dataset, under the assumption of working on Euclidean spaces as noted in section \ref{correctness}. This characteristic strongly limits the execution of complex instructions on CPUs to calculate square root or power values, and results in a valuable feature for the portability of the algorithm on embedded systems.
 
On the other hand, density-based clustering remains one of the most suitable approach for geospatial domains, and the linear computational cost in this context could also play an important role in the speed-up of processing on high-performance computing systems.

\subsection{Heuristic for parameter estimation}\label{heu}
\label{heuristic}
The discretization step used for the creation of the grid in \myalgocapitals is easily related to the input parameters of DBSCAN, $Eps$ and $MinPts$. 
This suggests that our algorithm may benefit from the same heuristics proposed for DBSCAN for an a-priori determination of the input parameters.

The method proposed in \cite{DBSCAN} for the estimation of $Eps$ and $MinPts$ is based on the analysis of the sorted \textit{k-dist} function plot.
Based on experimental results, it is stated that $MinPts = 4$ is an optimal choice for bi-dimensional datasets for DBSCAN.
Therefore the estimation of the parameter $Eps$ for DBSCAN can be evaluated from the observation of the 4-dist function plot $(k=MinPts)$.

We propose here a slightly modified version of this heuristic which has been proven to be a good tool for the estimation of $\gamma$ for \myalgocapitalsnospace.

As for DBSCAN, the first step consists in determining an estimate for $Eps$ from the direct observation of the sorted 4-dist plot. 

Let $Eps_{DBSCAN}$ be the chosen value; the relationship between $\gamma$ and $Eps$ given in section \ref{discretization_step} produces an immediate value for $\gamma$.
Nevertheless, to take into account inaccuracies due to space discretization, it is advisable to relax the constraint for the determination of an optimal value for $\gamma$ adopting the following criterion:

\begin{equation}
	\gamma \geq \frac{Eps_{DBSCAN}}{2\sqrt{2}}
\end{equation}
\null

\begin{figure}[H]
	\centering
	\fbox{
	\includegraphics[width=0.7\textwidth]{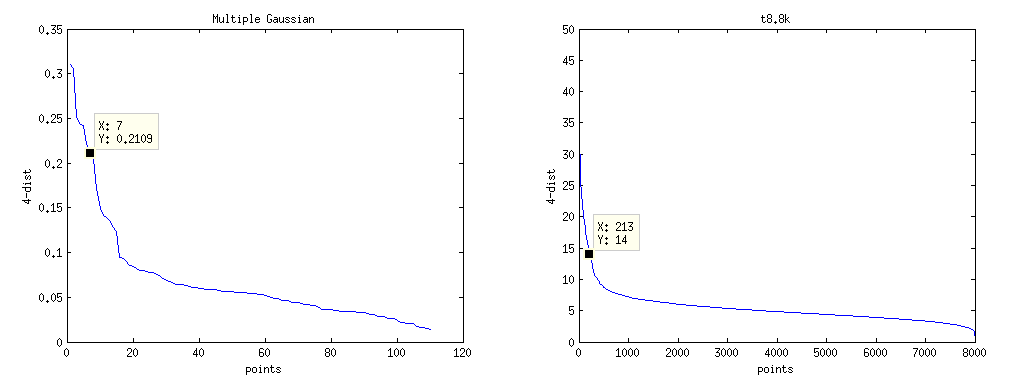}
	}
	\caption{K-dist plot for the Multiple Gaussians 2D and t8.8k datasets ($K = 4$). The black dot shows the value of $Eps > Eps_{DBSCAN}$ used to derive $\gamma$}
	\label{fig:heuplot}
\end{figure}

Figure \ref{fig:heuplot} shows the \textit{4-dist} plots generated for two of the testing datasets listed in table \ref{tab:dataset_table}. The value of $Eps$ corresponding to the optimal value of $\gamma$ is marked with a black dot, while the optimal value for DBSCAN, $Eps_{DBSCAN}$, is located approximately at the first ``valley''.

\medskip

Finally, given the ratio between the area of an $Eps-Neighborhood$ of DBSCAN and the area of cell in \myalgocapitalsnospace, we found that $MinPts=1$ is an optimal value for bi-dimensional datasets.

\subsection{Parameter sensitivity}
The sensitivity of \myalgocapitals to the variation of $\gamma$ and $MinPts$ is highly predictable.
Varying the value suggested for $\gamma$ by the heuristic previously described, can either merge clusters with close boundaries, or split a single cluster in multiple ones. 
 
\begin{figure}[H]
	\centering
	\frame{\includegraphics[width=0.7\textwidth]{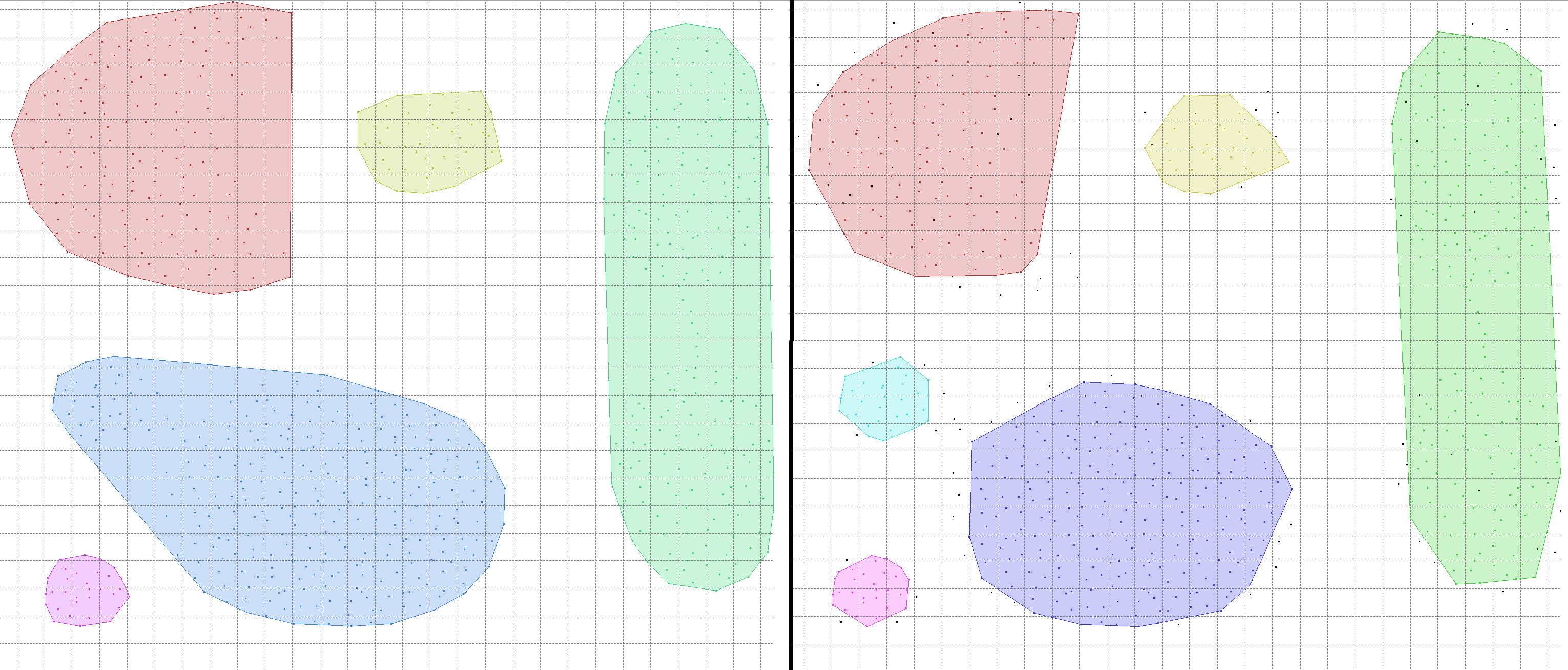}}
	\caption{Effect of increasing the parameter $MinPts$, keeping a fixed $\gamma$ value of $1.2$. On the left $MinPts = 1$, while on the right: $MinPts = 2$. }
	\label{fig:parameters}
\end{figure}

Nevertheless, keeping the size of the cells fixed, a similar behavior can be observed when the parameter $MinPts$ is changed.
Since $MinPts$ is used to determine whether a cell must be part of the current cluster or not, increasing its value can cause an erosion effect on the boundary of the clusters, as shown in figure \ref{fig:parameters}.
In some circumstances, this side effect can prevent the fusion of distinct clusters that share a low-density boundary area.
On the other hand, in case of non isodensity datasets, setting $MinPts > 1$ could imply the removal of the sparse inner cells in a connected region, thus producing ``hole'' in the clusters. 

Investigating the sentitivity of \myalgocapitals to the perturbation of the input parameters, we found out another significant difference between our approximated density model and DBSCAN.

The $MinPts$ threshold is for DBSCAN the desidered minimum size of a cluster, but, since the output of DBSCAN can slightly change scanning the input points in a different order, it my happen that clusters with less than $MinPts$ points are detected.
 
For \myalgocapitalsnospace, $MinPts$ is exactly the minimum number of points in a cluster, as previously stated in section \ref{fill_function}. Therefore, setting $MinPts = 1$, as suggested by the our modified heuristic method, will inhibit \myalgocapitals from discarding outliers, returning a single-point cluster for each noise point.

This drawback can easily be overcome by a filter procedure that removes all the clusters with a single element and, since this added algorithmic step has linear cost depending on the number of detected clusters, it does not invalidate the asymptotic analysis of the computational complexity of \myalgocapitals discussed in section \ref{complexity}.

Another solution consists in diverging from the strict application of the heuristic method: fixing $MinPts>1$, a value for $\gamma$ close to double of the one suggested for $MinPts=1$ should be selected. This is, however, not an exact rule to determine the optimal parameters for \myalgocapitalsnospace.  

\begin{figure}[H]
	\centering
	\frame{\includegraphics[width=0.8\textwidth]{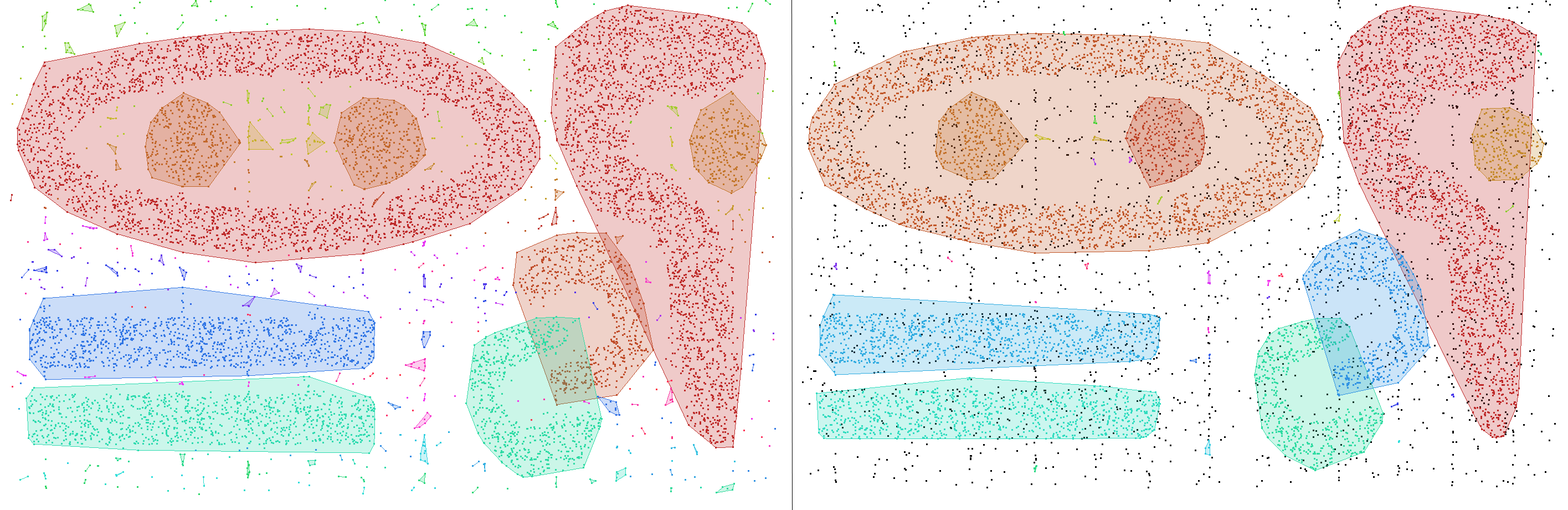}}
	\caption{Effect on noise detection increasing $\gamma$ and $MinPts$ on the dataset t7.10k. On the left $\gamma=4.31, MinPts=1$, on the right $\gamma=7.47, MinPts = 3$.}
	\label{fig:noiseminpts}
\end{figure}

Figure \ref{fig:noiseminpts} shows the similarity on clustering results obtained running \myalgocapitals with different values for $\gamma$ and $MinPts$ on the dataset \textit{t7.10k}. The convex hull of the elements in each cluster enhances the visualization of the cluster boundaries.
The first pair of input parameters was derived from the application of our heuristic method. The second pair, where $MinPts = 3$ produces as a side effect, a better detection of outliers and therefore, automatic removal of noise points (the black dots) from the output of \myalgocapitalsnospace. 

\section{Validation of clustering results}
\label{validation}
This section presents the experimental results obtained by \myalgocapitals on several benchmarking datasets.
We will focus on the evaluation of the clustering results on synthetic bidimensional datasets that have been used to test various validation techniques on the output produced by different clustering algorithms: \textit{Aggregation} \cite{aggregation}, \textit{Mouse}, \textit{Multiple Gaussians 2D} \cite{MultiGauss2D}, \textit{Pathbased} and \textit{Spiral} \cite{pathbased}, and \textit{Vary density} \cite{varydensity}. 
These 2D datasets provide a human-validated classification of the points and are available at the Internet links reported in table \ref{tab:dataset_table}. 


Given the availability of a ground truth for these datasets, we based our analysis on the use of several external validation indices. When compared to alternative evaluation methods, e.g. internal or relative indices that focus on the analysis of compactness and separability of clusters \cite{internalMeasures}, the external measures provide indeed an objective comparison of the quality of the results while are not biased by the shape of the clusters and are not sensitive to noise. Therefore, they results the best choice for the evaluation of density-based methods, while internal and relative measures are mostly suitable for the evaluation of partitional clustering algorithms, e.g. K-Means. 

Altough internal \cite{CDbw1} \cite{CDbw2} and relative \cite{DBVC} measures targeted to the evaluation of density-based clustering algorithms were developed, they provide results that are inherently fluctuating and sensitive to the parameters used to calculate the measure itself, therefore they are preferable to external indices only when no ground truth is available.

An interesting survey on the most used evaluation mothods and the related indices calculations can be found in \cite{CompIntExt} and \cite{ClusteringValidation}.

\begin{table} [h]
\caption{List of 2D datasets used to assess the performance of the algorithm \myalgocapitals \label{tab:dataset_table}}
 \begin{tabular}{|p{2.5cm}|r|p{9.5cm}|}
\hline 
\textbf{Name} & \textbf{Points} & \textbf{Source} \\ 
\hline 
Aggregation & 788  & http://cs.joensuu.fi/sipu/datasets/Aggregation.txt \\ 
\hline 
Pathbased & 300 & http://cs.joensuu.fi/sipu/datasets/pathbased.txt \\ 
\hline 
Spiral & 312 & https://cs.joensuu.fi/sipu/datasets/spiral.txt \\
\hline 
Mouse & 500 & http://elki.dbs.ifi.lmu.de/datasets/mouse.csv \\ 
\hline 
Vary density & 150 & http://elki.dbs.ifi.lmu.de/datasets/vary-density.csv \\ 
\hline 
{Multiple \linebreak Gaussian 2D} & 110 & {http://elki.dbs.ifi.lmu.de/browser/elki/trunk/data/ \linebreak synthetic/LoOP-publication/multiple-gaussian-2d/multiple-gaussian-2d.csv} \\ 
\hline 
t4.8k & 8000 & {http://glaros.dtc.umn.edu/gkhome/fetch/sw/cluto/ \linebreak chameleon-data.tar.gz} \\ 
\hline 
t5.8k & 8000 & {http://glaros.dtc.umn.edu/gkhome/fetch/sw/cluto/ \linebreak chameleon-data.tar.gz} \\ 
\hline 
t7.10k & 10000 & {http://glaros.dtc.umn.edu/gkhome/fetch/sw/cluto/ \linebreak chameleon-data.tar.gz} \\ 
\hline 
t8.8k & 8000 & {http://glaros.dtc.umn.edu/gkhome/fetch/sw/cluto/ \linebreak chameleon-data.tar.gz} \\ 
\hline 
\end{tabular} 
\end{table}

We used the heuristics suggested in \cite{DBSCAN} to determine the proper parametrization for DBSCAN on each of these datasets, and tried to maximize the score for each index while preserving the qualitative result of the clustering.
Then we applied our modified heuristic method based on the estimates of the $Eps_{DBSCAN}$ values to derive a suitable set of parameters for \myalgocapitalsnospace, that are shown in table \ref{tab:paramstable}.
Table \ref{tab:indici} reports the results obtained for DBSCAN and \myalgocapitals with this set of parameters.

\begin{table}[H]
\caption{Parameters estimated with the heuristic methods for \myalgocapitals \label{tab:paramstable} and DBSCAN}
\begin{center}
\begin{tabular}{l|l|c|l|c}
 & \multicolumn{2}{|c}{\textbf{DBSCAN}} & \multicolumn{2}{|c}{\textbf{\myalgocapitals}}\\  
 & \multicolumn{1}{|c|}{\scriptsize{\textbf{$Eps$}}} & \scriptsize{\textbf{$MinPts$}} & \multicolumn{1}{|c|}{\scriptsize{\textbf{$\gamma$}}} & \scriptsize{\textbf{$MinPts$}} \\ 
\hline
\small{Aggregation} & 1 & 4 & 1.2 & 2 \\
\small{Mouse} & 0.035 & 4 & 0.02035 & 1\\
\small{Multiple Gaussians 2D} & 0.1 &  4 & 0.035 & 1\\
\small{Pathbased} & 1.5 & 4 & 0.8 & 1\\
\small{Vary density} & 0.0675 & 4& 0.03 & 1 \\
\small{Spiral} & 1.2 & 1 & 1 & 1 \\
\hline
\end{tabular}
\end{center}
\end{table}

\begin{table}[H]
\caption{Comparison by validation indices between DBSCAN and \myalgocapitals on heuristic-derived paramenters\label{tab:indici}}
\resizebox{\textwidth}{!}{
\begin{tabular}{l|c|c|c|c|c|c}
 & \multicolumn{2}{|c}{Aggregation} & \multicolumn{2}{|c}{Mouse} & \multicolumn{2}{|c}{Multiple gaussian}\\  
 & \textbf{DBSCAN} & \textbf{\myalgocapitals} & \textbf{DBSCAN} & \textbf{\myalgocapitals} & \textbf{DBSCAN} & \textbf{\myalgocapitals} \\ 
\hline
\small{Precision}   	& \textbf{0.8265}	& 0.7486 			& \textbf{0.9432} 	& 0.9313 & 0.9208 & \textbf{1.0000} \\
\small{Recall}      	& 0.9258 		  	& \textbf{1.0000} 	& 0.7434 & \textbf{0.7559} & \textbf{0.9359} & 0.7957 \\
\small{F-measure}   	& \textbf{0.8733} 	& 0.8562 			& 0.8314 & \textbf{0.8345} & \textbf{0.9588} & 0.8862 \\
\small{Rand}        	& \textbf{0.9416} 	& 0.9273 			& 0.8744 & \textbf{0.8753} & \textbf{0.9740} & 0.9354 \\
\small{Jaccard}     	& 0.7751 			& \textbf{0.7486} 	& 0.7115 & \textbf{0.7160} & \textbf{0.9208} & 0.7957 \\
\small{F.-Mallows}  	& \textbf{0.8747} 	& 0.8652 			& 0.8387 & \textbf{0.8390} & \textbf{0.9590} & 0.8920 \\
\small{NMI}		  	& 0.8876 			& \textbf{0.8949} 	& \textbf{0.7441} & 0.7276 & \textbf{0.9216} & 0.8324 \\
\hline
\multicolumn{7}{l}{} \\
 & \multicolumn{2}{|c}{Pathbased} & \multicolumn{2}{|c}{Vary density} & \multicolumn{2}{|c}{Spiral}\\ 
 & \textbf{DBSCAN} & \textbf{\myalgocapitals} & \textbf{DBSCAN} & \textbf{\myalgocapitals} & \textbf{DBSCAN} & \textbf{\myalgocapitals} \\ 
\hline
\small{Precision}   & 0.8945 & \textbf{0.9975} & 0.9887 & \textbf{1.0000} & 1.0000 & 1.0000 \\
\small{Recall}      & \textbf{0.5702} & 0.5262 & \textbf{0.7469} & 0.6519 & 1.0000 & 1.0000 \\
\small{F-measure}   & \textbf{0.6964} & 0.6890 & \textbf{0.8510} & 0.7893 & 1.0000 & 1.0000 \\
\small{Rand}        & 0.8334 & \textbf{0.8418} & \textbf{0.9128} & 0.8855 & 1.0000 & 1.0000 \\
\small{Jaccard}     & \textbf{0.5343} & 0.5256 & \textbf{0.7406} & 0.6519 & 1.0000 & 1.0000 \\
\small{F.-Mallows}  & 0.7142 & \textbf{0.7245} & \textbf{0.8594} & 0.8074 & 1.0000 & 1.0000 \\
\small{NMI}  		& 0.6190 & \textbf{0.6544} & \textbf{0.8162} & 0.7336 & 1.0000 & 1.0000 \\
\hline
\end{tabular}}
\end{table}

Coherently with the differences in the theoretical models used by \myalgocapitals and DBSCAN, some differences between the two algorithms can be observed. 
This is obviously the effect of the approximation in the discrete density model used by \myalgocapitalsnospace, that reduces the accuracy of classification results especially for the boundary elements of each cluster. 

Furthermore, as for DBSCAN, the heuristic applied to estimate suitable parameters for \myalgocapitals must be regarded as a rough indication to start the research of optimal values.
Following this line of reasoning, we have done further experiments aimed at the research of optimal parameters for \myalgocapitals on these same datasets, that are shown in table \ref{tab:optimal-paramstable}. The score achieved with these set of parameters is show in table \ref{tab:optimal-indici}.

\begin{table}[H]
\caption{Optimal parameters for \myalgocapitals \label{tab:optimal-paramstable}}
\begin{center}
\begin{tabular}{l|l|c|l|c}
 & \multicolumn{2}{|c}{\textbf{\myalgocapitals}}\\  
 & \multicolumn{1}{|c|}{\scriptsize{\textbf{$\gamma$}}} & \scriptsize{\textbf{$MinPts$}} \\ 
\hline
\small{Aggregation}  & 0.595 & 1 \\
\small{Mouse}  & 0.021 & 1\\
\small{Multiple Gaussians 2D} & 0.171 & 2\\
\small{Pathbased}  & 0.826 & 1\\
\small{Vary density} & 0.03615 & 1 \\
\small{Spiral} & 1 & 1 \\
\hline
\end{tabular}
\end{center}
\end{table}

In general, with the use of optimal parameters, \myalgocapitals outperforms DBSCAN in almost all the testing cases. 

\begin{table}[H]
\caption{Comparison by validation indices between DBSCAN and \myalgocapitals with optimal paramenters\label{tab:optimal-indici}}
\resizebox{\textwidth}{!}{
\begin{tabular}{l|c|c|c|c|c|c}
 & \multicolumn{2}{|c}{Aggregation} & \multicolumn{2}{|c}{Mouse} & \multicolumn{2}{|c}{Multiple gaussian}\\  
 & \textbf{DBSCAN} & \textbf{\myalgocapitals} & \textbf{DBSCAN} & \textbf{\myalgocapitals} & \textbf{DBSCAN} & \textbf{\myalgocapitals} \\ 
\hline
\small{Precision}   & 0.8265 & \textbf{0.8445} & 0.9432 & \textbf{0.9434} & 0.9208 & \textbf{0.9448} \\
\small{Recall}      & 0.9258 & \textbf{0.9568} & 0.7434 & \textbf{0.8413} & 0.9359 & \textbf{0.9763} \\
\small{F-measure}   & 0.8733 & \textbf{0.8971} & 0.8314 & \textbf{0.8894} & 0.9588 & \textbf{0.9603} \\
\small{Rand}        & 0.9416 & \textbf{0.9525} & 0.8744 & \textbf{0.9131} & 0.9740 & \textbf{0.9745} \\
\small{Jaccard}     & 0.7751 & \textbf{0.8134} & 0.7115 & \textbf{0.8009} & 0.9208 & \textbf{0.9236} \\
\small{F.-Mallows}  & 0.8747 & \textbf{0.8989} & 0.8387 & \textbf{0.8909} & 0.9590 & \textbf{0.9604} \\
\small{NMI}			& 0.8876 & \textbf{0.8998} & 0.7441 & \textbf{0.7879} & \textbf{1.0000} & 0.9641 \\
\hline
\multicolumn{7}{l}{} \\
 & \multicolumn{2}{|c}{Pathbased} & \multicolumn{2}{|c}{Vary density} & \multicolumn{2}{|c}{Spiral}\\ 
 & \textbf{DBSCAN} & \textbf{\myalgocapitals} & \textbf{DBSCAN} & \textbf{\myalgocapitals} & \textbf{DBSCAN} & \textbf{\myalgocapitals} \\ 
\hline
\small{Precision}   & 0.8921 & \textbf{0.9899} & 0.9887 & \textbf{1.0000} & 1.0000 & 1.0000 \\
\small{Recall}      & 0.5722 & \textbf{0.5920} & \textbf{0.7469} & 0.7069 & 1.0000 & 1.0000 \\
\small{F-measure}   & 0.6972 & \textbf{0.7409} & \textbf{0.8510} & 0.8283 & 1.0000 & 1.0000 \\
\small{Rand}        & 0.8334 & \textbf{0.8600} & \textbf{0.9128} & 0.9036 & 1.0000 & 1.0000 \\
\small{Jaccard}     & 0.5351 & \textbf{0.5885} & \textbf{0.7406} & 0.7069 & 1.0000 & 1.0000 \\
\small{F.-Mallows}  & 0.7144 & \textbf{0.7655} & \textbf{0.8594} & 0.8408 & 1.0000 & 1.0000 \\
\small{NMI}  	 	& 0.6190 & \textbf{0.6967} & \textbf{0.8162} & 0.7725 & 1.0000 & 1.0000 \\
\hline
\end{tabular}}
\end{table} 

The only case where \myalgocapitals exhibits lower efficacy than DBSCAN is with the dataset \textit{Vary-density}. In this peculiar case, the presence of clusters with variable densities exposes some limitations of the approximated density model used by \myalgocapitalsnospace. Therefore, all the validation indexes, with the only exception of the precision, are lower than those achieved by DBSCAN. 

Nevertheless, it is also worth noting that the difference between the scores reported by the two algorithms is bound to no more than 4\%, and in general the sensitivity of the approximated model to space discretization is more evident with datasets with low cardinalities, as in the case of testing synthetic datasets.

As a final consideration, the remarkable speedup of \myalgocapitals in terms of performance enables the possibility to derive a procedure for the automatic determination of optimal parameters based on the ones derived by the heuristic.  

Therefore, from this analysis, \myalgocapitals appears to be a resilient alternative to DBSCAN for all the applications whose effectiveness is bound to performance.   

\subsection{Results with the chameleon dataset}
In order to further evaluate the algorithm here presented, \myalgocapitals has been tested also with the bi-dimensional datasets \textit{t4.8k}, \textit{t5.8k}, \textit{t7.10k} and \textit{t8.8k} created for the algorithm Chameleon \cite{Chameleon}, available at the URL shown in table \ref{tab:dataset_table}. 

As explained by the authors, these datasets were generated with no predefined model with reference to density, shape, similarity, and so on. As a consequence they represent a challenge to clustering algorithms in general, and it will be shown that good clustering results can be achieved by \myalgocapitals with a proper parametrization.

\begin{figure} [H]
\centering
\fbox{\includegraphics[width=1\textwidth]{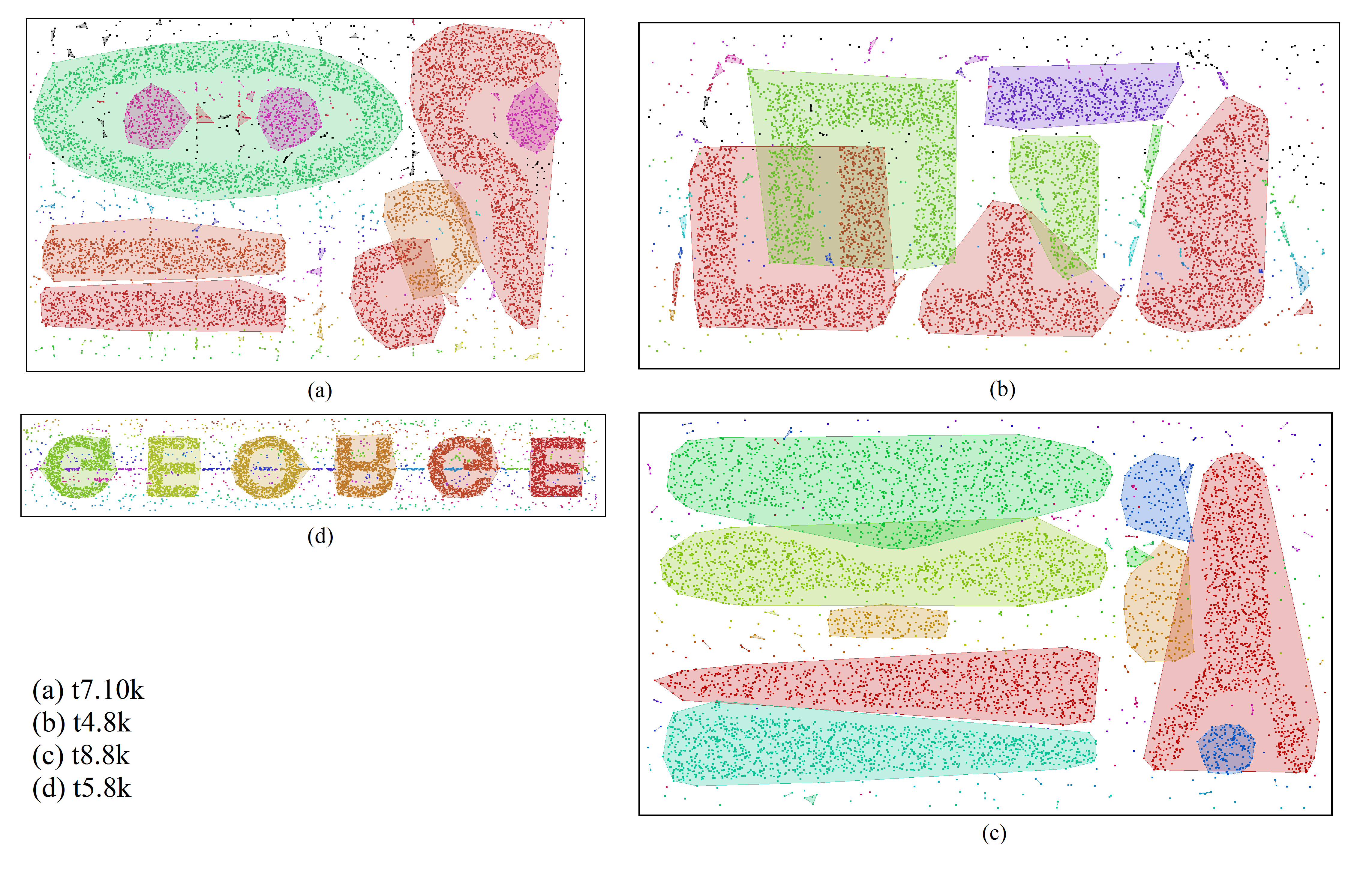}}
\caption{Ouput of \myalgocapitals on the Chameleon datasets; a convex hull encloses the elements of each cluster.} \label{fig:cham}
\end{figure}

Figures \ref{fig:cham} show the clusters found by \myalgocapitals for the dataset \textit{t4.8k}, \textit{t5.8k}, \textit{t7.10k} and \textit{t8.8k} respectively.

It is interesting to note about the latter dataset that, coherently with the limitations of the density models of both DBSCAN and our algorithm, all the clusters, except the one with lower density, are correctly detected by \myalgocapitalsnospace.

Table \ref{tab:paramstable-chameleon} shows the parameters used on these datasets derived from the application of our heuristic method.

\begin{table}[H]
\caption{Parameters used for \myalgocapitals on the Chameleon datasets \label{tab:paramstable-chameleon}}
\begin{center}
\begin{tabular}{l|l|c|l|c}
  & \multicolumn{2}{|c}{\textbf{\myalgocapitals}}\\  
  & \multicolumn{1}{|c|}{\scriptsize{\textbf{$\gamma$}}} &\scriptsize{\textbf{$MinPts$}} \\ 
 \hline
 \small{t4.8k}   & 3.34  & 1 \\
 \small{t5.8k}   & 1.873 & 1 \\
 \small{t7.10k}  & 4.31  & 1 \\
 \small{t8.8k}   & 4.94  & 1 \\
 \hline
\end{tabular}
\end{center}
\end{table}

\section{Experimental results}
\label{experiments}
We investigated the dependency of the execution time varying both the number of points and the dimensionality of the input dataset.

The computer used for this experiments is equipped with an Intel processor (i7-4790K @ 4.00 GHz) and a 16.0 GB RAM memory. The operating system used was GNU/Linux 64-bit.

In order to highlight the speed-up factor of \myalgocapitals over DBSCAN we produced synthetic bi-dimensional datasets 50k, 100k, 200k, 500k and 1M that were populated with random points. 

Table \ref{tab:tempi} reports the execution times of an $O(n \log{n})$ implementation of DBSCAN, that makes use of KD-tree indexing structures, and \myalgocapitals on the Chameleon datasets a our synthetic datasets.

\begin{table}[ht]
\caption{Execution times in milliseconds varying the number of points.\label{tab:tempi}}
\resizebox{\textwidth}{!}{
\begin{tabular}{r|r|r|r|r|r|r|r|r}
\textbf{Dataset} & \textbf{Points} & \textbf{DBSCAN} & \textbf{$Eps$} & \textbf{$MinPts$} & \textbf{\myalgocapitals} & \textbf{$\gamma$} & \textbf{$MinPts$} & \textbf{Speed-up} \\
\hline 
1M      & 1000000   & 1385965 & 11.28 & 10  & 3893 & 4     & 10 & 356 \\
500k 	  & 500000  	& 413684  & 16.92 & 14	& 1640 & 6     & 14 & 252 \\
200k 	  & 200000  	& 46985 	& 18.33 & 10	& 619  & 6.5   & 10 & 75 \\
100k 	  & 200000  	& 10151 	& 20.00 & 3	  & 350  & 2.5   & 3  & 29 \\
50k 	  & 200000  	& 3930		& 35.25 & 3   & 83   & 12.5  & 3  & 47 \\
\hline
t7.10k 	& 10000  		& 112		 & 12.15 & 1	  & 37 	 & 4.31  & 1   & 3.0 \\
t5.8k 	& 8000  		&  89		 & 5.28  & 1	  & 38 	 & 1.873 & 1  & 2.3 \\
t4.8k 	& 8000  		&  79		 & 9.41  & 1	  & 36 	 & 3.34  & 1   & 2.3 \\
t8.8k 	& 8000  		&  83 	 & 13.93 & 1	  & 24 	 & 4.94  & 1   & 3.3 \\
\hline
\end{tabular}}
\end{table}

\begin{figure}[h!]
	\centering
	\fbox{
	\includegraphics[width=1\textwidth]{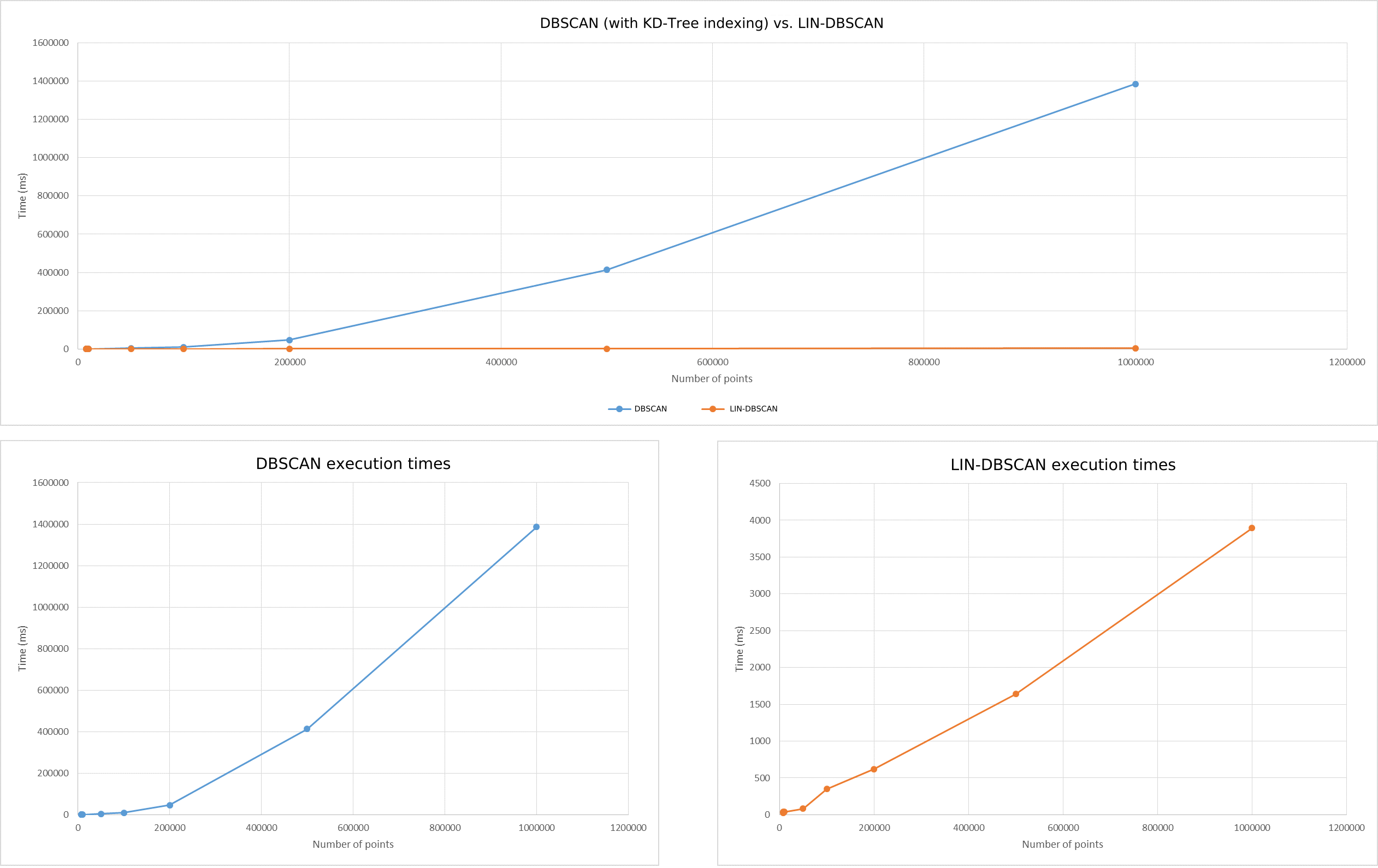}
	}
	\caption{Relation between the execution time in milliseconds and the number of points}
	\label{fig:output}
\end{figure}

Figure \ref{fig:output} show the escalation of DBSCAN execution times compared to the almost linear scaling of \myalgocapitals with reference to the cardinality of the dataset. The time needed to build the KD-tree for DBSCAN and the grid hash table for \myalgocapitals are included in the measurements.


Table \ref{tab:tempi-dimensioni} shows the relationship between execution times and the dimensionality of input dataset.

\begin{table}[H]
\caption{Execution times in ms varying with the dimensionality of the dataset.\label{tab:tempi-dimensioni}}
\begin{center}
\begin{tabular}{c|c|c|c}
\textbf{Dataset} & \textbf{Dimensions} & \textbf{Points} & \small{\textbf{\myalgocapitals}} \\
\hline
150k-6D & 6 	& 150000	& 1484 \\
150k-5D & 5 	& 150000	& 1102 \\
150k-4D & 4 	& 150000	& 845  \\
150k-3D & 3 	& 150000	& 687  \\
150k-2D & 2 	& 150000	& 429  \\
\hline
\end{tabular}
\end{center}
\end{table}

The curse of dimensionality is a well known problem in data analysis. DBSCAN and \myalgocapitals make no exception. Increasing the dimensionality of the input dataset will negatively affect the performance of both algorithms. We tested this phenomena using synthetic bi-dimensional datasets of 12k points with increasing dimensionality.

In DBSCAN the increase of execution time depends mainly on the metrics used to compute the distances between the input data. For the euclidean distance, for instance, each extra dimension implies a sum, a subtraction, and a multiplication. 

In \myalgocapitals the dimensionality affects the number of accesses to the grid. The reference implementation used to produce the experimental results here discussed is based on a hash map. By increasing the number of dimensions, the cardinality of the set of the keys becomes higher than the cardinality of the set of indices. In this case, a collision (two keys corresponding to a unique index) is more likely, and the access cost for a cell is no longer constant and this results in a substantial increase in run time.
Therefore \myalgocapitalsnospace, with a higher number of dimensions, clearly exhibits an exponential growth in execution times when the order of magnitude of $3^d$ approaches $n$.

\section{Parallel implementation}
\label{parallel}
One of the main advantages of grid-based clustering techniques is the possibility of easy parallelization. There are several ways this can be achieved. 
For example, it is possible to execute in a concurrent context of the \emph{fill} procedure making each thread start from a different cell in the grid. If two threads reach the same cell during their scan, the two partial clusters are merged and only one of the two threads will continue to run. Listing \ref{algo:scanparallel} shows the pseudo-code for this parallel implementation of the \emph{fill} function.

Synchronization is achieved using the function \emph{synchronizeAccess()} which acts like a synchronization barrier. One way to implement it is using a mutex for each non empty cell in the grid. However, this may cause an excessive memory usage, depending on the size of the input data set.

Another possible solution to write a parallel version of \myalgocapitals is to subdivide the grid into sub grids (i.e. 4 or 16) making each thread work on one sub grid only. When each thread terminates the execution of \myalgocapitalsnospace, a simple procedure can analyze the boundary cells of adjacent sub grids, to determine whether or not two clusters identified in different sub grids need to be merged.

\SetKwProg{FnParFill}{Function}{}{return}
\begin{algorithm}[H]
\SetAlgorithmName{Listing}{}{}
  \FnParFill{fill($cluster$, $cell$, $MinPts$)}
  {
    syncrhonizeAccess($cell$)\;
    \If{$cell$ is not empty}
    {
	    \If{$cell$.numPoints $>= MinPts$}
	    {
		    \If{$cell$ is not visited}
		    {
			    $cell$.visited = true\;
			    \ForEach{p in cell.points}
			    {
				    $cluster$.addPoint(p) \;
			    }			

			    $neighborCells$ = findNeighbors($cell$)\;
			    \ForEach{$neighbor$ in $neighborCells$}
			    {
				    fill($cluster$, $neighbor$, $MinPts$)\;
			    }
		    }
		    \Else
		    {
			    mergeClusters($cluster$, $cell$.cluster)\;
			    \Return{}
		    }
	    }
    }
  }
  \caption{Parallel version of FILL}
  \label{algo:scanparallel}
\end{algorithm}

\section{Applications}
\label{application}
In the recent past, a considerable interest has been growing for the adoption of density based clustering algorithms in computer vision.
The main reason can be found in the commercial diffusion of cheap, yet effective, depth sensing devices that can reproduce a 3D representation of the environment within a given range and resolution in the form of points cloud.

One of the most used is the Kinect sensor, that besides its commercial use as innovative input device for entertainment platforms, has gained popularity in scientific research thanks to the availability of open source resources like the OpenNI framework \cite{openNI}.

The research efforts have been mainly devoted to the integration of this sensor technology in real-time robotic motion, shape recognition and understanding. 
Each of this tasks starts with the analysis of the points cloud produced by the sensor and because of this, most of the times, clustering is one of the early steps in the process \cite{kinect-legs-detection}  \cite{kinect-body-measurment}.
Furthermore, given the peculiarities of this domain (low dimensionality, noise presence, and uniform sampling of the environment) density based clustering is the most suitable technique.

In order to prove the effectiveness of \myalgocapitalsnospace, we have selected an existing application that applies a clustering techniques in the field of computer vision. In \cite{kinect-legs-detection} the authors present a framework for the detection of human legs in motion from a low perspective. This result is achieved with the integration of the Kinect sensor in the motion control system of the quadruped robot StarlETH \cite{Hutter_1starleth} to provide a collision avoidance functionality.

The authors also provide an on-line reference to the source code package developed to fulfill the task. A C++ implementation of DBSCAN is used as a pre-processing step in order to identify distinct points clouds to use as input to the legs detection module, implemented with a set of Matlab routines.

We have integrated our reference C++ implementation of \myalgocapitals into this package in order to perform a comparison between DBSCAN and \myalgocapitals in terms of efficiency. The results of our tests are shown in the table below.
Apart from the substitution of a single line of code in order to invoke the \textit{linDBSCANClustering} routine in place of DBSCAN, no other parts of the original software package were modified for the execution of these tests.

\begin{figure} [H]
\centering
\fbox{\includegraphics[width=0.7\textwidth]{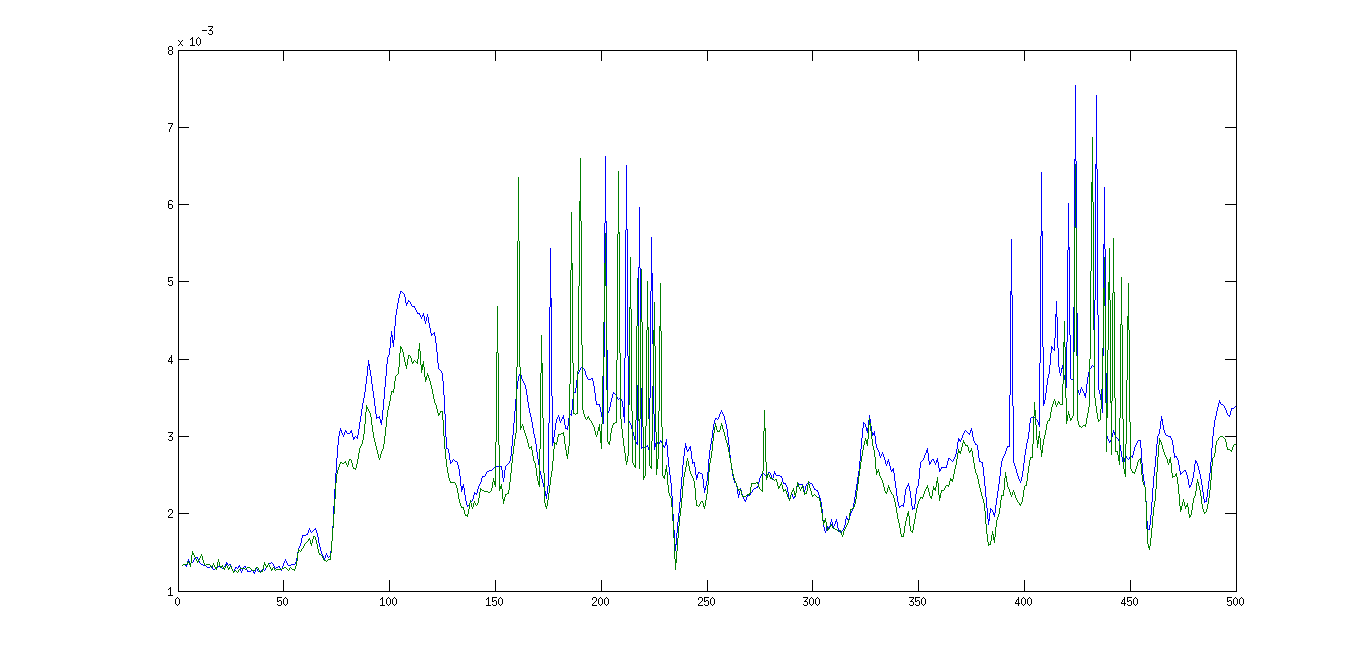}}
\caption{Comparison between DBSCAN (blue) and \myalgocapitals (green) in terms of execution time per each frame on the S-Easy dataset} 
\label{fig:s-easy-times}
\end{figure} 
 
\begin{figure} [H]
\centering
\fbox{\includegraphics[width=0.7\textwidth]{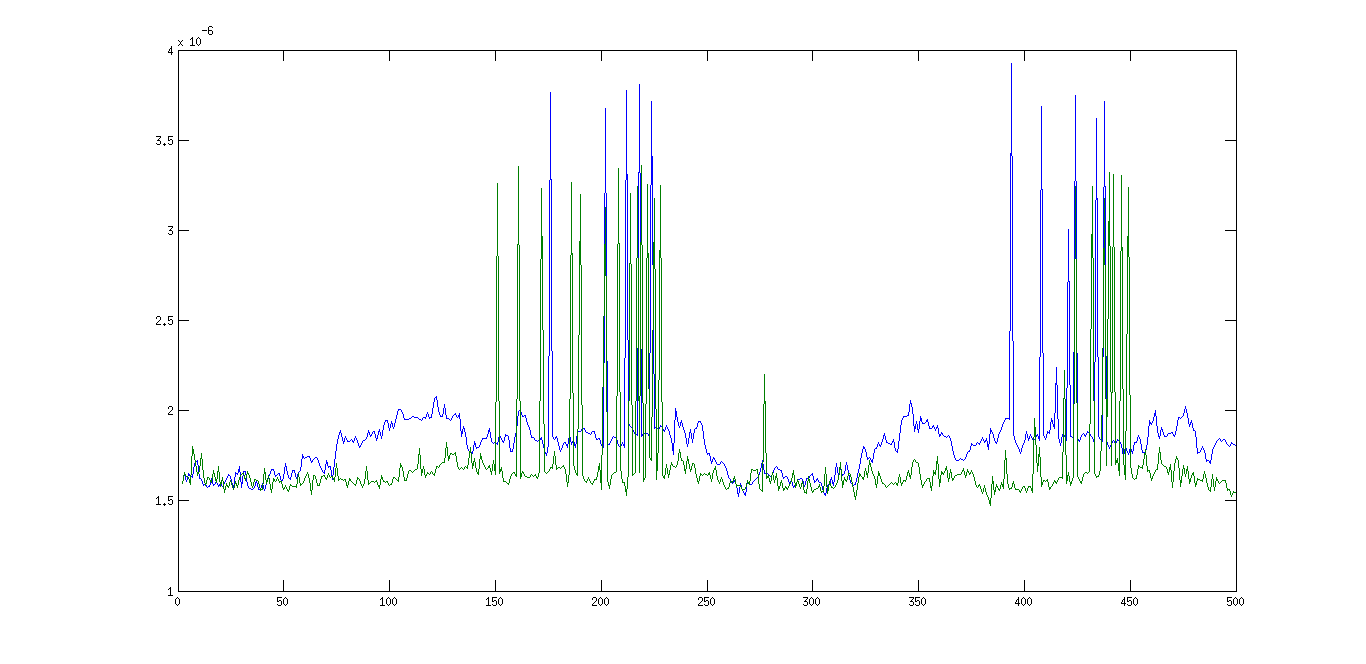}}
\caption{Comparison between DBSCAN (blue) and \myalgocapitals (green) in terms of execution time and number of points ratio per each frame on the S-Easy dataset} \label{fig:s-easy-times-vs-points}
\end{figure} 

The tests were performed using one of the testing dataset provided by the authors in OpenNI (oni) file format. The execution times of the application running the two algorithms on an Intel i7 processor are reported in terms of FPS in the table \ref{tab:kinect-ffdbc-dbscan} while figures \ref{fig:s-easy-times} and \ref{fig:s-easy-times-vs-points} show respectively the execution times and the ratio between execution time and number of points for DBSCAN (in blue) and \myalgocapitals (in green) on the dataset S-Easy.
 
It can be seen that, even if the clustering step is only a part of the overall process, an increment in terms of performance has been achieved with \myalgocapitalsnospace, preserving the collision avoidance functionality.

The time measurements were taken with the use of the \textit{tic} and \textit{toc} API provided by Matlab. 
It is possible to observe several spikes in both the execution times of DBSCAN and \myalgocapitals due to the state of the operating system (context switching, memory swap, etc.). All these factors have to be taken into account because they can affect the precision of the measurements especially in this case, where the execution times of both the algorithms are in the order of milliseconds. Those spurious effects would become less evident when increasing the size of the input arrays, for instance using a depth sensor with higher resolution, and the performance gap between the two algorithms would be even more evident. 

Nevertheless it is possible to observe a uniform difference between the times achieved by DBSCAN and \myalgocapitals, both in a comparison based on a frame by frame basis and in the average times values. 
  
\begin{table}[H]
\caption{Comparison of DBSCAN and \myalgocapitals performance on the S-Easy dataset\label{tab:kinect-ffdbc-dbscan}}
\begin{tabular}{c|c|c|c|c}
 \textbf{Algorithm} & \textbf{Avg FPS} & \textbf{Max FPS} & \textbf{Min FPS} & \textbf{Avg Clustering time} \\ 
 \hline
 \myalgocapitals & 18.39 & 107.38 & 9.12 & 0.0025 s\\
 DBSCAN & 17.49 & 41.83 & 9.40 & 0.0027 s\\
 \hline
\end{tabular}
\end{table}
 
The source code package modified to integrate \myalgocapitals in the leg detection chain is available at the following URL: \url{https://git.io/v2EFC}.

\section{Conclusion and Future Works}
\label{conclusions}
This paper presents a new density-based clustering algorithm, \myalgoextendednospace, that features a linear time complexity. This result is achieved thanks to the adoption of a discrete version of the density model first introduced with DBSCAN.

Experimental results prove that, even if \myalgocapitals provides a considerable increase in efficiency, its clustering results are almost identical to those produced by DBSCAN in terms of quality.

Since its appearance, DBSCAN has gained a significant role in the panorama of clustering algorithms. 
In fact, in comparison to other well known methods and algorithms, it is based on a simple, yet solid, model that ties the concept of cluster to the density feature only, without imposing constraints on shape or distribution. 

This is the reason why so much effort has been devoted to the research of alternative methods and improvements that could solve its major drawbacks: high computational cost, varying density and the curse of dimensionality.

As for the original DBSCAN algorithm, the possibility to adapt this \myalgocapitals to datasets with high dimensionality is bound to the definition of a metric that can fit in the density model.

Nevertheless, clustering does not apply only in the \textit{big data} field, where high dimensionality is a relevant issue. There are, indeed, many applications that can benefit of the reduced computational costs of \myalgocapitals on datasets that are tailored for the density-based clustering. 

As an example, several real-time computer vision techniques can take advantage of this kind of fast clustering techniques in their processing chain.
This is also true for pattern recognition techniques that need to manipulate raw input with noise. 
 
Given its linear computational cost, \myalgocapitals could represent a valid option for these applications, especially when they need to cope with high performance requirements.

Therefore, in order to expand the capabilities of this technique, our efforts are currently oriented to the elaboration of a version of \myalgocapitals that is able to cope with varied density, preserving the improvements in performance.

\bibliographystyle{plain}
\bibliography{bibliografia.bib}

\end{document}